\documentclass[10pt,twocolumn,letterpaper]{article}

\usepackage{cvpr}
\usepackage{times}
\usepackage{epsfig}
\usepackage{graphicx}
\usepackage{amsmath}
\usepackage{amssymb}
\usepackage{multirow}

\cvprfinalcopy % *** Uncomment this line for the final submission

\def\cvprPaperID{***} % *** Enter the CVPR Paper ID here

\begin{document}

%%%%%%%%% TITLE
\title{Attention-aware Multi-stroke Style Transfer}

\author{Yuan Yao$^1$,~Jianqiang Ren$^2$,~Xuansong Xie$^2$,~Weidong Liu$^1$,~Yong-Jin Liu$^1$,~Jun Wang$^3$\\
$^1$Tsinghua University, ~~~~$^2$Alibaba Group, ~~~~$^3$University College London\\
}

\maketitle
%\thispagestyle{empty}

%%%%%%%%% ABSTRACT
\begin{abstract}

Neural style transfer has drawn considerable attention from both academic and industrial field. Although visual effect and efficiency have been significantly improved, existing methods are unable to coordinate spatial distribution of visual attention between the content image and stylized image, or render diverse level of detail via different brush strokes. In this paper, we tackle these limitations by developing an attention-aware multi-stroke style transfer model. We first propose to assemble self-attention mechanism into a style-agnostic reconstruction autoencoder framework, from which the attention map of a content image can be derived. By performing multi-scale style swap on content features and style features, we produce multiple feature maps reflecting different stroke patterns. A flexible fusion strategy is further presented to incorporate the salient characteristics from the attention map, which allows integrating multiple stroke patterns into different spatial regions of the output image harmoniously. We demonstrate the effectiveness of our method, as well as generate comparable stylized images with multiple stroke patterns against the state-of-the-art methods.

\end{abstract}

%%%%%%%%% BODY TEXT
\section{Introduction}
\label{sec:Introduction}

Style transfer is a powerful technique for art creation and image editing that enables recomposing the images in the style of other images. Recently, inspired by the power of Convolutional Neural Network (CNN) in visual perception tasks, Gatys \etal~\cite{gatys2015texture,gatys2016image} opens up a new field named Neural Style Transfer, which first introduces neural representations to separate and recombine content and style of arbitrary images. They propose to extract the content features and style correlations (Gram matrix) along the processing hierarchy of a pretrained network classifier. Based on this work, several algorithms have been proposed to accelerate the development in terms of the generalization and efficiency issues, through optimization-based methods~\cite{gatys2016image} and feed-forward networks~\cite{johnson2016perceptual,li2016precomputed,ulyanov2016texture}. The success of style transfer makes it possible to deploy service upon mobile applications (e.g., Prisma, Artify), allowing users to create an artwork out of a picture they took with their phones.

\begin{figure}[t]
  \centering
  \resizebox{0.98\linewidth}{!}{
   \includegraphics{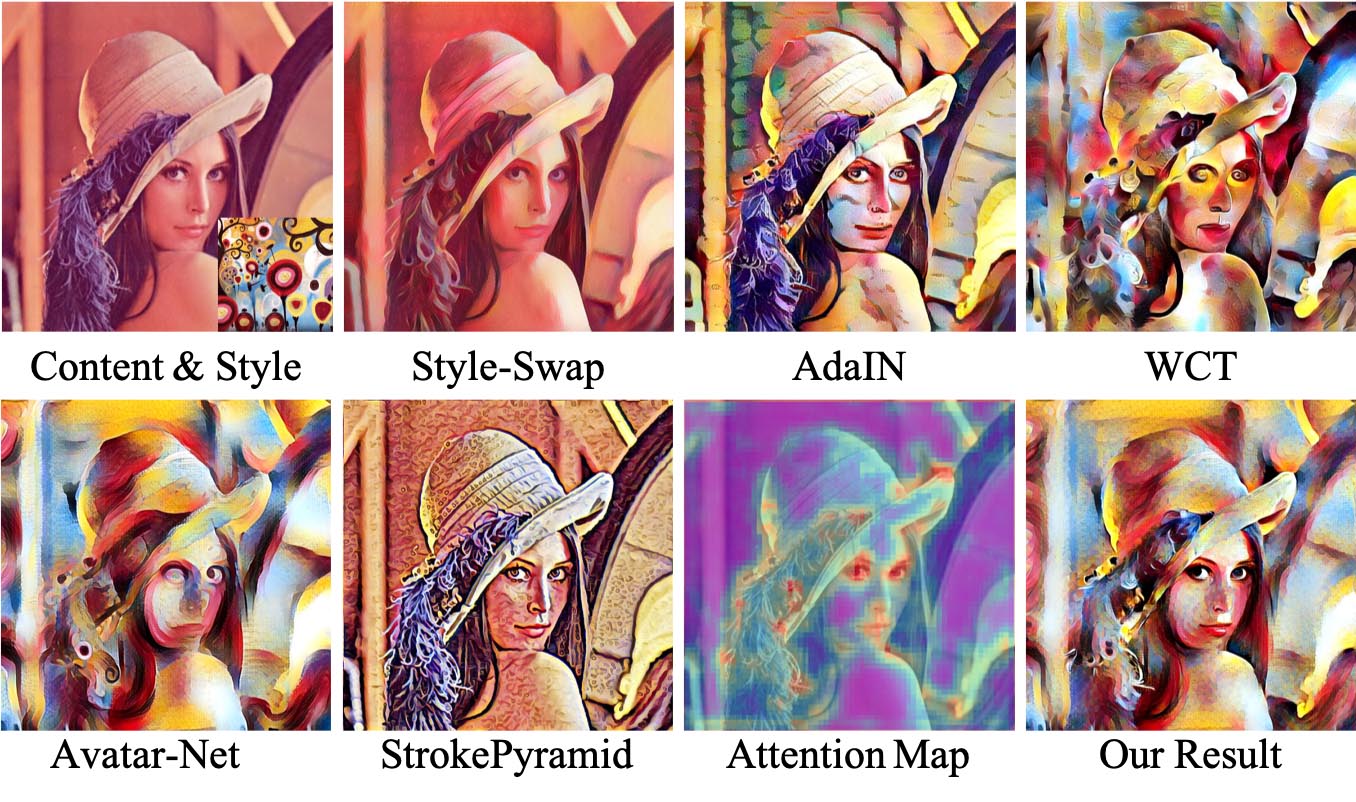} }
  \caption{Existing methods are unable to coordinate spatial distribution of visual attention between content image and stylized image, misleading the stylized rendering on diverse content regions in an indiscriminate way, resulting in insufficient stylization (Style-Swap), unexpected patterns (AdaIN, StrokePyramid) or distorted attention regions (WCT, Avatar-Net). Due to attention mechanism and multi-stroke fusion strategy, our method can achieve faithful transfer of style and attention consistency with content image simultaneously.}
  \label{fig: motivation}
  \vspace{-15pt}
\end{figure}

In spite of significant progress already achieved, these methods suffer from the restricted binding between the model and specific styles. Recently, {\em Arbitrary-Style-Per-Model} Fast Style Transfer Methods (ASPM)~\cite{jing2017neural} are proposed to conquer this dilemma. One possible solution is to coordinate high-level statistical distribution between content features and style features. Although visual quality and efficiency can be greatly improved, they unexpectedly introduce unexpected or distorted patterns to the stylized result because of treating diverse image regions in an indiscriminate way, such as AdaIN~\cite{huang2017arbitrary} and WCT~\cite{li2017universal} in Figure~\ref{fig: motivation}. Another solution is to swap the content feature patch with the closest style feature patch at the intermediate layer of a trained autoencoder. However, this method may generate insufficient stylized results when huge difference exists between content and style images, such as the Style-Swap~\cite{chen2016fast} method shown in Figure~\ref{fig: motivation}. Compared with Style-Swap, Avatar-Net~\cite{sheng2018avatar} further dispels the domain gap between content and style features, leading to better stylized results, but it still maintains inconsistent spatial distribution of visual attention with the content image and thus manifests distortion in terms of semantic perception.

Stroke textons~\cite{zhu2005textons}, which are referred to fundamental micro-structures in natural images, reflect perceptual style patterns. Methods such as~\cite{jing2018stroke,wang2017multimodal} are dedicated to learn stroke control in transfer process. Jing \etal~\cite{jing2018stroke} first proposes to achieve continuous stroke size control by incorporating multiple stroke sizes into one single StrokePyramid model. The StrokePyramid result shown in Figure~\ref{fig: motivation} is produced by mixing two different stroke sizes. However, due to the lack of local awareness of content image, they perform the stroke interpolation in a holistic way regardless of region diversity, leading to lacking of level of details. In addition, these methods are inflexible to handle arbitrary styles in one feedforward pass.

To address the aforementioned problems, we propose an attention-aware multi-stroke (AAMS) model for arbitrary styles transfer. Our model encourages {\em attention consistency} (which refers to spatial consistency of visual attention distribution) for corresponding regions between content image and stylized image, and it achieves both scalable multi-stroke fusion control and automatic spatial stroke size control in one shot. Specifically, we introduce self-attention mechanism as complementary to the autoencoder framework. The self-attention module calculates the response at a position as a weighted sum of the features at all positions, which helps to capture long-range dependencies across image regions. By performing a reconstruction training process for the self-attention assembled autoencoder, the attention map could grasp salient characteristics within any content images. As shown in Figure~\ref{fig: motivation}, the attention map of the content image highlights the salient parts while enabling consistency of the attention degree for long-range features. Based on the correlation between receptive field and stroke size, a multi-scale style swap module is proposed to blend distinct stroke patterns via swapping the content features with multi-scale style features in high-level representations. We inject the attention map into a multi-stroke fusion module to synthesize distinct stroke patterns harmoniously, which achieves automatic spatial stroke size control. Comprehensive experiments have been conducted to demonstrate the effectiveness of our method, and the model is capable of generating comparable stylized images with multiple stroke patterns against the state-of-the-art methods. The main contributions of this work are:
\begin{itemize}
\item We introduce self-attention mechanism to an autoencoder network, allowing capturing critical characteristics and long-range region relations of the input image.
\item We propose multi-scale style swap to break the limitation of fixed receptive field in high-level feature space and produce multiple feature maps reflecting different stroke patterns.
\item By combining with attention map, we present a flexible fusion strategy to integrate multiple stroke patterns into different spatial regions of the output image harmoniously, which enables the attention consistency between the content image and stylized image. 
\end{itemize}

\begin{figure*}[ht]
  \centering
  \resizebox{0.98\linewidth}{!}{
   \includegraphics{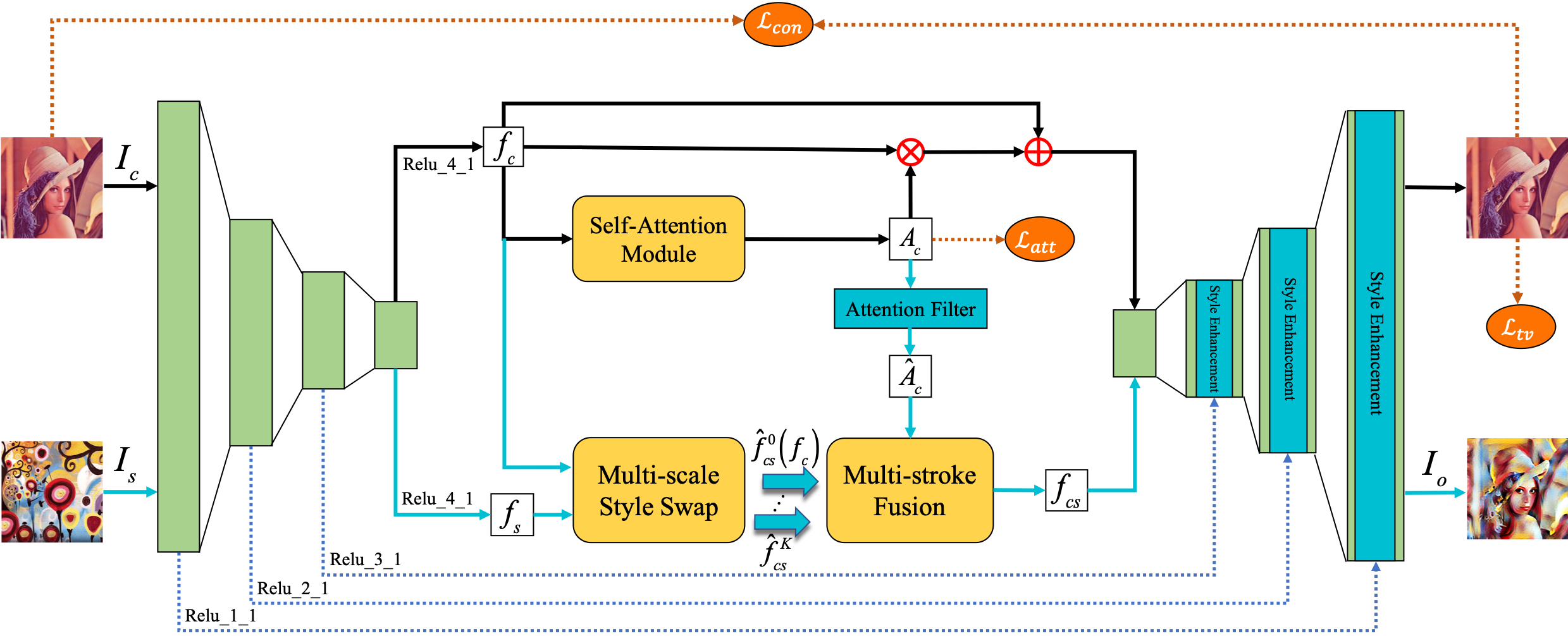} }
  \caption{An overview of our network architecture. We train a self-attention autoencoder for image reconstruction in a style-agnostic manner. Given a content image $I_{c}$ and a style image $I_{s}$ in the testing phase, we perform multi-stroke style transfer via style swap and fusion modules under the guidance of attention map $\hat{A}_{c}$. Skip connections are introduced to enhance the stylization effects during the reconstruction process. The black and green lines represent the training and testing phases respectively, while the orange dotted line denotes the flow of our loss functions.}
  \label{fig: architecture}
  \vspace{-10pt}
\end{figure*} 

%-------------------------------------------------------------------------

\section{Related Work}
\label{sec: Related Work}

\noindent {\bf Neural Style Transfer.} Arbitrary-Style-Per-Model methods (ASPM)~\cite{jing2017neural} have recently been proposed to transfer arbitrary styles through only one single model. The backend idea is to formulate the style transfer as an image reconstruction process, with feature statistics fusion between content and style features at intermediate layers. Chen \etal~\cite{chen2016fast} first proposes to swap the content feature patch with the best matching style feature patch using a Style-Swap operation. Huang \etal~\cite{huang2017arbitrary} introduces adaptive instance normalization (AdaIN) to adjust the mean and variance of the content feature to match those of the style feature. Li \etal~\cite{li2017universal} performs the style transfer by integrating the whitening and coloring transforms (WCT) to match the statistical distributions and correlations between the features of content and style. Avatar-Net~\cite{sheng2018avatar} elevates the feature transfer ability by matching the normalized counterparts features and applying a patch-based style decorator. However, the above methods either locally exchange the closest feature patches or transfer feature statistics globally, which tends to display a uniform stroke pattern without attention aware guaranty. In comparison, our methods perform style transfer to render visually plausible result with multi-stroke patterns integrated within the same stylized image. Another related work is~\cite{jing2018stroke}, which proposes a StrokePyramid module to incorporate multiple stroke sizes into one single model, it empowers distinct stroke sizes in different spatial regions within the same transformed image. Our approach can automatically manipulate multiple stroke fusion through the guidance of attention feature in one shot within ASPM framework, while~\cite{jing2018stroke} achieves spatial stroke size control by feeding masked content image and need to be retrained for each new style.

\noindent {\bf Attention Models.} One of the most promising trends in research is the emergence of incorporating attention mechanism into deep learning framework~\cite{larochelle2010learning,mnih2014recurrent}. Rather than compressing an entire image or a sequence into a static representation, attention allows the model to focus on the most relevant part of images or features as needed. Such mechanism has been proved to be very effective in many vision tasks including image classification~\cite{xiao2015application,zhou2016learning}, image captioning~\cite{xu2015show,you2016image} and visual question answering~\cite{xu2016ask,yang2016stacked}. In particular, self-attention~\cite{lin2017a,vaswani2017attention} has been proposed to calculate the response at a position in a sequence by attending to all positions within the same sequence. Shaw \etal~\cite{shaw2018self} proposes to incorporate relative position information for sequences in the self-attention mechanism of the Transformer model, which improves the translation quality on machine translation tasks. Zhang \etal~\cite{zhang2018self} demonstrates that the self-attention model can capture the multi-level dependencies across image regions and draw fine details in the context of GAN framework. Compared with \cite{zhang2018self}, we adapt the self-attention to introduce a residual feature map to catch salient characteristics within content images.

%-------------------------------------------------------------------------

\section{Proposed Approach}
\label{sec: Proposed Approach}

For the task of style transfer, the goal is to generate a stylized image $I_{o}$, given the content image $I_{c}$ and style image $I_{s}$. To satisfy arbitrary style transfer in one feed-forward pass while integrating images with multiple stroke patterns, we develop three modules( {\em multi-scale style swap}, {\em multi-stroke fusion} and {\em self-attention module}) in the bottleneck layer for feature manipulation. The three modules cooperate with each other and form a coupling framework. The network architecture of our proposed approach is depicted in Figure~\ref{fig: architecture}.

Assume that $f_{c}$ and $f_{s}$ denote the feature maps extracted from the encoder correspond to $I_{c}$ and $I_{s}$ respectively. At the core of our network, a {\em self-attention module} is proposed to learn the saliency properties by feeding with $f_{c}$, which will generate self-attention feature map $\emph{A}_{c}$ for any content $I_{c}$ after reconstruction training process. During the intermediate process of transferring $f_{c}$ into the domain of $f_{s}$ via a WCT transformation in the testing phase, a {\em multi-scale style swap module} is first designed to synthesize features of multiple stroke sizes $\hat{f}_{cs}^{k}(k=1,\ldots, K)$, taking both the content feature $f_{c}$ and style features $f_{s}$ as input. The module conducts a style-swap procedure for the content feature and \emph{K} style features simultaneously. To perform a flexible integration of the features, a {\em multi-stroke fusion module} is presented to handle controllable blending. The attention map $\hat{A}_{c}$ filtered from $\emph{A}_{c}$ is incorporated to guide fusion among the content feature $f_{c}(\hat{f}_{cs}^{0})$ and the $K$ swapped stroke features $\hat{f}_{cs}^{k}(k=1,\ldots, K)$, where $K$ is the user provided clustering number. After the two steps, the synthetic feature ${f}_{cs}$ is fed into the trained decoder to generate the stylized image $I_{o}$ in one feed-forward pass. We further introduce skip connections to enhance the stylization effects by adapting multiple level of synthetic features with style features. In this section, we introduce the three components in details.

%-------------------------------------------------------------------------

\subsection{Self-Attention Autoencoder}
\label{sec: Self-Attention Autoencoder}

We employ a self-attention mechanism~\cite{zhang2018self} to catch the relationships between separated spatial regions. In detail, let $\Phi_{e}(\cdot)$ and $\Phi_{d}(\cdot)$ represent the encoder and decoder respectively. The encoder produces $f_{x} = \Phi_{e}(x) \in \mathbb{R}^{H \times W \times C}$ by mapping the input image \emph{x} into the intermediate activation space, with \emph{H}, \emph{W}, \emph{C} denoting height, width and channel number respectively. 

Let $flat(\cdot)$ denote the flatten operation along the channel dimension. We calculate the self-attention feature of $f_{x}$ in the $i^{th}$ location as:

\begin{equation} \label{eq: SA}
A_{x}^{i} = \sum_{j=1}^{HW} \alpha_{ij} \ast flat(f_{x} \otimes \Theta_{h})_{j}\,,
\end{equation}
where $\otimes$ denotes the convolution operation and $\alpha_{ij}$ is the weight coefficient. $\alpha_{ij}$ indicates the dependencies between two regions, not just for neighboring positions. It is computed using a softmax function:

\begin{equation} \label{eq: SA_weight}
\alpha_{ij} = \frac{\exp(e_{ij})}{\sum_{k=1}^{HW} \exp(e_{ik})}\,,
\end{equation}
where $e$ is obtained by a compatibility function that compares elements by a scaled dot product:

\begin{equation} \label{eq: SA_dot}
e = flat(f_{x} \otimes \Theta_{u}) \ast flat(f_{x} \otimes \Theta_{g})^{T}\,,
\end{equation}

The self-attention feature map decoded from the self-attention module is $\emph{A}_{x} = (a_{x}^{1}, a_{x}^{2},\ldots,a_{x}^{N})^{T} \in \mathbb{R}^{N \times C}, N = HW$, which will be reshaped as the same dimensions of $f_{x}$. In the above formulation, $\Theta_{h}, \Theta_{u}, \Theta_{g} \in \mathbb{R}^{C\times \hat{C}}$ are learned parameter matrices within our self-attention autoencoder framework. We implement them as 1 $\times$ 1 convolution~\cite{zhang2018self}.

Given the self-attention feature map $\emph{A}_{x}$, we propose to obtain a self-attention residual $R_{x}$ by multiplying the hidden feature map $f_{x}$ with $\emph{A}_{x}$. The output is then given by appending the residual to the feature map $f_{x}$.

\begin{equation} \label{eq: SA_output}
O_x= R_{x} + f_{x} = A_{x} \odot f_{x} + f_{x}\,,
\end{equation}
where $\odot$ denotes the element-wise multiplication operator. We then feed $O_{x}$ into the decoder and reconstruct the input image $\hat{x} = \Phi_{d}(O_{x})$. In this manner, the self-attention residual $R_{x}$ can exhibit the salient regions in synthesizing the image while charactering the correlations between distant regions. 

Similar to~\cite{johnson2016perceptual,li2017universal,sheng2018avatar}, we define the semantic content loss as the sum of perceptual loss and pixel reconstruction loss to generate visually indistinguishable images with the input images:

\begin{equation} \label{eq: content function}
\mathcal{L}_{con} = \sum_{l\in l_c}||\phi_{l}(\hat{x}) - \phi_{l}(x)||_{2}^{2} + \lambda_{p}||\hat{x} - x||_{2}^{2}\,, 
\end{equation}
where $\phi_{l}(x)$ is the activations of the $\emph{l}^{th}$ layer of the pretrained VGG-19 network when processing image \emph{x}, $\lambda_{p}$ is the weight to balance the two losses. Both losses are calculated using normalized Euclidean distance. In addition, we introduce a sparse loss on self-attention feature map $A_x$ to encourage the self-attention autoencoder to pay more attention to small regions instead of the whole image:

\begin{equation} \label{eq: attention loss}
\mathcal{L}_{att} = ||A_{x}||_{1}\,.
\end{equation}

With total variational regularization loss $\mathcal{L}_{tv}$~\cite{johnson2016perceptual} added to encourage the spatial smoothness in the generated images, we obtain our total loss function as:

\begin{equation} \label{eq: overall loss}
\mathcal{L} = \lambda_{con}\mathcal{L}_{con}  + \lambda_{att}\mathcal{L}_{att} + \lambda_{tv}\mathcal{L}_{tv}\,,
\end{equation}
where $\lambda_{con}$, $\lambda_{att}$ and $\lambda_{tv}$ are the balancing factors.

%-------------------------------------------------------------------------

\subsection{Multi-scale Style Swap}
\label{sec: Multi-scale Style Swap}

Style swap is the process of substituting content feature with the closest style feature patch-by-patch~\cite{chen2016fast}. Given a specific patch size, the style swap procedure can be implemented efficiently as two convolutions operations and a channel-wise argmax operation. The filters of the convolutional layers are derived from the extracted style patches. Based on the analysis of how receptive field influences stroke size in~\cite{jing2018stroke}, we note that larger patch size leads to larger receptive field and style stroke accordingly. However, the increase of the scale for patch size is strictly limited by the network structure and easily saturated when the patch size is larger than the fixed receptive field of network.

To resolve the above issue in an efficient manner, we propose to fix the patch size while changing the scale of activation feature map of styles, such that we introduce multi-scale stroke patterns after swapping with the same content feature. Specifically, we first adopt whitening transform~\cite{li2017universal} on content feature $f_{c}$ and style feature $f_{s}$ to peel off their style information while preserving global structure, results are denoted as $\hat{f}_{c}$  and $\hat{f}_{s}$. We then obtain a series of multi-scale style features by casting the whitened style feature into multiple scales:
                  
\begin{equation} \label{eq: multi-scale style feature}
\hat{f}_{s}^{k} = \mathcal{T}_{\beta_{k}}(\hat{f}_{s})\,,
\end{equation}
where $\mathcal{T}$ denotes the scaling operation, $\beta_{k} (k = 1,2,\ldots,K)$ is the scale coefficient controlling different stroke sizes. Finally, the multiple swapped features are produced by performing style-swap procedure between $\hat{f}_{c}$ and multiple $\hat{f}_{s}^{k}$ simultaneously.

\begin{equation} \label{eq: multi-scale style swap}
\hat{f}_{cs}^{k} = \mathcal{F}_{ss}(\hat{f}_{c}, \hat{f}_{s}^{k})\,,
\end{equation}
where $\mathcal{F}_{ss}$ denotes the parallelizable style swap process~\cite{chen2016fast}. 

\begin{figure}[t]
  \centering
  \resizebox{0.98\linewidth}{!}{
   \includegraphics{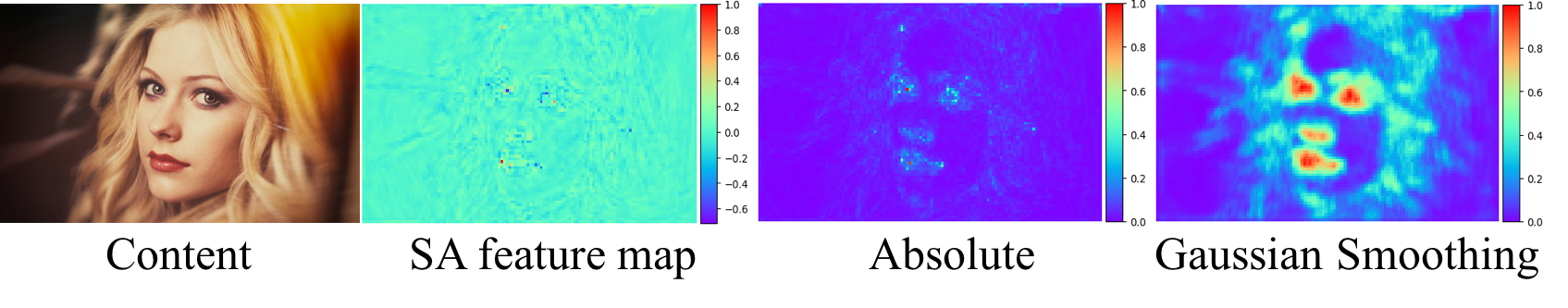} }
   % \vspace{-5pt}
  \caption{\emph{Attention filter} conducts Absolute and Gaussian smoothing transformation upon the self-attention feature map, the generated attention map can exhibit evident saliency distribution.}
  \label{fig: attention_map}
  \vspace{-10pt}
\end{figure}

%-------------------------------------------------------------------------
\subsection{Multi-stroke Fusion}
\label{sec: Multi-stroke Fusion}
Equipped with self-attention mechanism, the residual $R_{c}$ in eq.~\eqref{eq: SA_output} is able to capture critical characteristics and long-range region correlations of the feature map $f_{c}$. In detail, during the reconstruction process, the residual learns to fine tune intrinsically crucial parts of content feature by adding variation to it, therefore the non-trivial (zero) parts of the residual deserve special attention. We utilize an \emph{attention filter} by first performing an absolute operation to highlight these non-trivial parts in $A_{c}$, following by a Gaussian kernel convolution layer to enhance the regional homogeneity of the features. The variance of the Gaussian kernel can further be used to control the salient region's proportion in content image. We obtain our \emph{attention map} $\hat{A}_{c}$ after normalizing into the range $[0,1]$. Figure~\ref{fig: attention_map} shows the visualization image of the intermediate results. We notice that the attention map can enlarge the attending influence of saliency regions while maintaining the correlations among distant regions. 

In addition to the \emph{K} swapped features $\hat{f}_{cs}^{k}$ obtained from the multi-scale style swap, we introduce the whitened content feature $\hat{f}_{c}$ as another entry to character the most significant regions, which is known as the fine-grained stroke $\hat{f}_{cs}^{0}$. Therefore, we have in total of \emph{K}+1 features for multi-stroke fusion.

To integrate arbitrary strokes in a scalable framework, we propose a flexible fusion strategy by first dividing the attention map $\hat{A}_{c}$ into multiple clusters according to the stroke number provided. We apply k-means method to cluster our attention map, the goal is to iteratively find \emph{K}+1 intensity centers, minimizing the Euclidean distance between center and elements among each clusters.

\begin{equation} \label{eq: kmeans}
\arg \min \mathcal{G} = \sum_{k=0}^{K}\sum_{\hat{a}_{c}^{i} \in S_{k}}||\hat{a}_{c}^{i} - m_{k}||^2\,,
\end{equation}
where \emph{K}+1 clusters are generated, and $m_{k} \in [0,1]$ denotes the mean intensity value of all the attention points $\hat{a}_{c}^{i}$ in cluster $S_{k}$. 

The multi-stroke fusion can then be formulated as integrating the content feature with multiple swapped features under the guidance of attention map.

\begin{equation} \label{eq: Multi-stroke Fusion}
\hat{f}_{cs} = \sum_{k=0}^{K} \hat{A}_{c}^{k} \hat{f}_{cs}^{k}\,,
\end{equation}
where $\hat{A}_{c}^{k} $ is the weight map assigned to the $k^{th}$ stroke size, and \emph{k} = 0 denotes the fine-grained stroke. The weight map is computed by a softmax function:

\begin{equation} \label{eq: smoothing weight}
\hat{A}_{c}^{k} = \frac{e^{\gamma \cdot (1 - D_{k})}}{\sum_{i=0}^{K}e^{\gamma \cdot (1 - D_{i})}}\,.
\end{equation}
We define $D_{k} = |\hat{A}_{c} - m_{k}|$ to measure the absolute distance with regard to the center $m_{k}$, thus $1-D_{k}$ can be used to indicate how much extent each stroke size contributes to synthesis the feature. The smoothing factor $\gamma$ is used to control the smoothing degree for fusion.

Before feeding into the decoder to generate the stylized result, we derive $f_{cs}$ by performing coloring transform with the syncretic feature $\hat{f}_{cs}$ to match feature statistics to the style feature $f_{s}$ following~\cite{li2017universal}.

%-------------------------------------------------------------------------

\begin{figure*}[t]
  \centering
  \resizebox{0.98\linewidth}{!}{
   \includegraphics{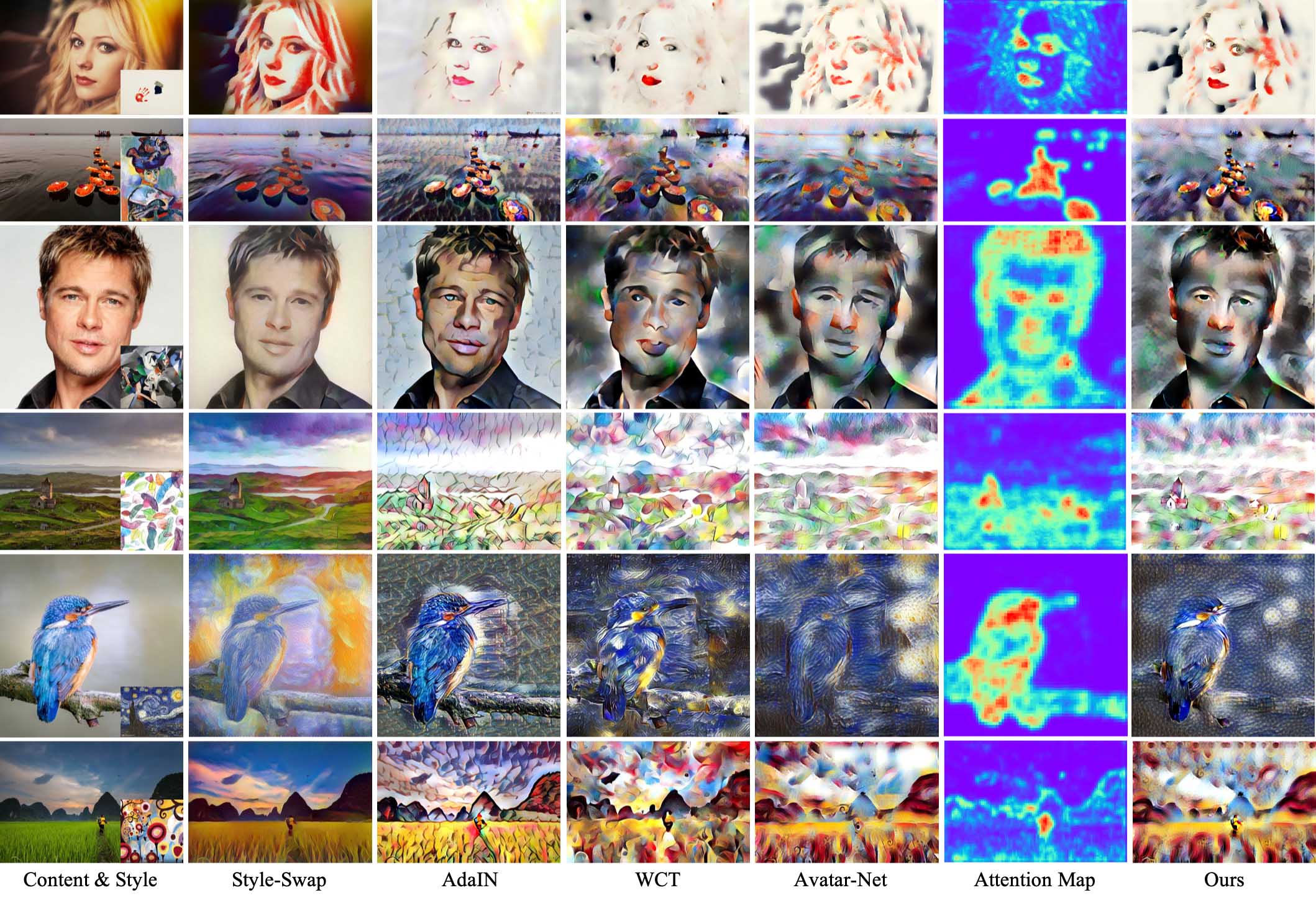} }
   % \vspace{-5pt}
  \caption{Results from different ASPM style transfer methods. We evaluate various content images including portraits, animals, landscapes with distinctive styles. The parameters of the prior methods are set as the default values according to their papers. More results will be presented in the supplementary material.}
  \label{fig: experiment_cmp}
  \vspace{-10pt}
\end{figure*}

\section{Experiments}
\label{sec: Experiments}

\subsection{Implementation Details}

We train our self-attention autoencoder for reconstruction using MS-COCO dataset~\cite{lin2014microsoft}, which contains roughly 80,000 training examples. We preserve the aspect ratio of image and rescale the smaller dimension to 512 pixels, and then randomly crop to 256 $\times$ 256 pixels. Our encoder contains the first few layers from VGG-19 model~\cite{simonyan2014very} pre-trained on the ImageNet dataset~\cite{deng2009imagenet}. The decoder is symmetric to the encoder structure. The Relu\_X\_1 (X = 1, 2, 3, 4) in the encoder are used to compute the perceptual loss and $\lambda_{con}, \lambda_{p}, \lambda_{att}, \lambda_{tv}$ are set as 1, 10, 6, 10 to balance each loss. We use three 1 $\times$ 1 convolutions for $\Theta_{h}, \Theta_{u}, \Theta_{g}$ in our self-attention module, and set $\hat{C} = C$ for $\Theta_{h}$ and $C/2$ for $\Theta_{u}, \Theta_{g}$. We train our network using the Adam optimizer and a batch size of 8 for 10k iterations. Although our method can handle arbitrary number of stroke fusions, we select three stroke scenario as a default setting in the following experiments, including the fine-grained stroke size and two coarse stroke sizes $(K=2)$ whose scale coefficients $\beta_{k}$ are 0.5 and 1 respectively. We set the smoothing factor $\gamma$ as 50 to reveal discriminative patterns for different stroke.

To obtain abundant stylized information, it is advantageous to match features across different levels in the VGG-19 encoder to fully capture the characteristics of the style as suggested in~\cite{li2017universal,sheng2018avatar}. We hence adopt a similar strategy for our reconstruction process. We use skip connections to perform style enhancement using adaptive instance normalization~\cite{huang2017arbitrary}, feeding with style features extracted from Relu\_X\_1 (X = 1, 2, 3) outputs and the reconstruction features in corresponding deconvolution layers.

%-------------------------------------------------------------------------

\subsection{Qualitative Evaluation}
\noindent {\bf Comparison With Prior Methods.} We evaluate four state-of-the-art methods for arbitrary style transfer: Style-Swap~\cite{chen2016fast}, AdaIN~\cite{huang2017arbitrary}, WCT~\cite{li2017universal} and Avatar-Net~\cite{sheng2018avatar}. The stylized results for various content/style pairs are shown in Figure~\ref{fig: experiment_cmp}. To make a fair comparison, the results of compared methods are obtained by running their codes with default configurations. Style-Swap simply relies on the patch similarity between content and style feature, so as to strictly preserve the content feature, which is validated in its results that only low-level style patterns (e.g., colors) are transferred. AdaIN presents an efficient solution by directly adjusts the content feature to match the mean and variance of the style feature. But it usually brings similar texture patterns for all the stylized images (e.g., the crack pattern in all stylized results) due to the style-dependent training. WCT holistically adjusts the feature covariance of content feature to match that of style feature, which inevitably brings unseen information and unconstrained patterns, for example the missing circular patterns in the $6^{th}$ row and the unexpected textures in the $5^{th}$ row. Avatar-Net shrinks the domain gap between content and style features and enhances propagation for feature alignment. Although concrete style patterns are reflected, it still cannot handle attention-aware feature adaption and manifest distortion from semantically perception, for example the eyes on the first row and the farmer on the last row. 

By contrast, our method can produce visually plausible stylized results against the previous methods. The attention map enables the seamlessly synthesis among multiple stroke sizes, while demonstrating superior spatial consistency of visual attention between content image and stylized image. In the last column of Figure~\ref{fig: experiment_cmp}, the salient regions in content images such as eyes, candles, house and farmer still maintain the focus in the stylized image. This validates the effectiveness of the attention map in the $6^{th}$ column, where salient regions are mainly attributes to the fine-grained stroke size. In addition, the attention map exhibits similarity measurement between distant regions, which allows detailed features in distant portions of the image are consistent with each other. \medskip 

\begin{figure}[t]
  \centering
  \resizebox{0.98\linewidth}{!}{
   \includegraphics{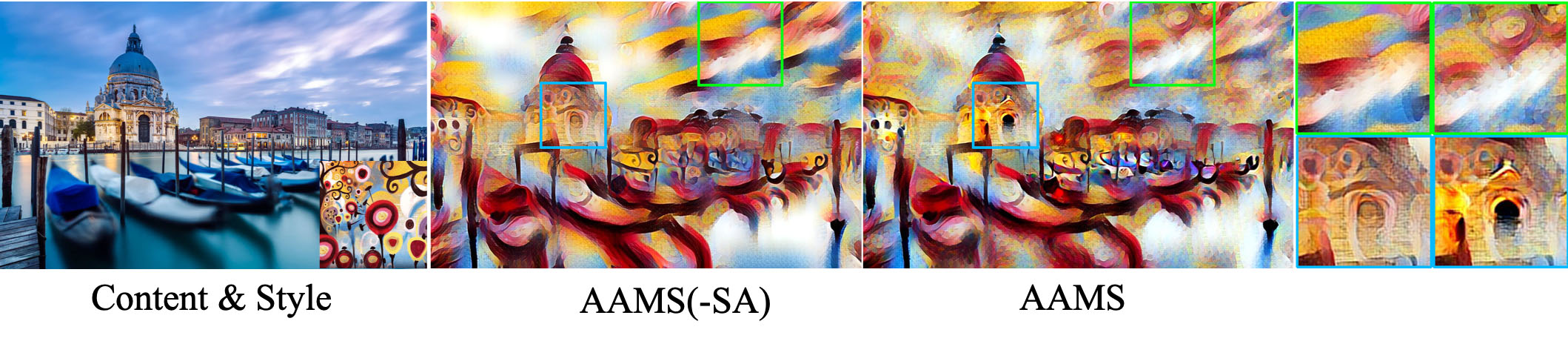} }
   % \vspace{-10pt}
  \caption{Comparison results to present the effectiveness of attention guidance. Regions marked by bounding boxes are zoomed in for a better visualization.}
  \label{fig: ablation}
\end{figure}

\begin{table}[t]
  \centering\scriptsize
  % \vspace{-5pt}
  \resizebox{1\linewidth}{!}{
  \begin{tabular}{c|ccccc}
 \hline
 Method & AUC\_Judd $\uparrow$ & SIM $\uparrow$ & NSS $\uparrow$ & CC $\uparrow$ & KL $\downarrow$ \\
 \hline
 AAMS(-SA) & 0.479 & 0.677 & 3.627 & 0.732 & 0.458 \\
 AAMS & 0.484 & 0.744 & 4.172 & 0.834 & 0.396\\ 
 \hline
\end{tabular}}
  \vspace{5pt}
  \caption{Quantitative measurements on saliency maps predicted by SalGAN~\cite{Pan_2017_SalGAN}. The attention consistency is dramatically improved with our method for all the evaluation metrics~\cite{bylinskii2018different}.}
  \label{tab:consistency}   
  \vspace{-10pt}
\end{table}

\begin{figure*}[t]
  \centering
  \resizebox{0.9\linewidth}{!}{
   \includegraphics{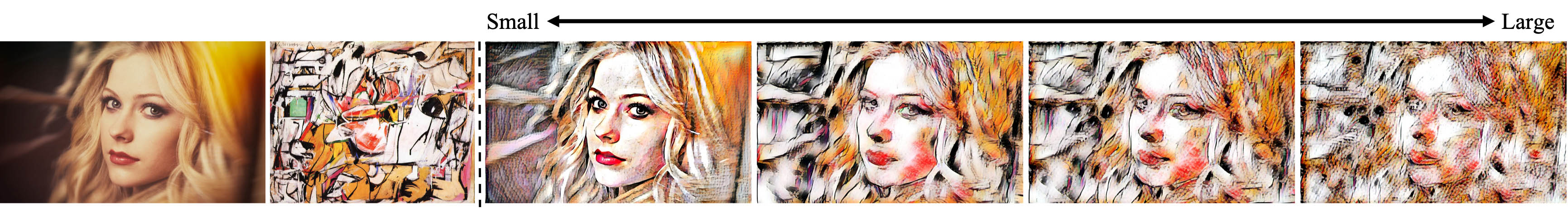} }
  \vspace{-3pt}
  \caption{Stylized result via different stroke sizes.}
  \label{fig: stroke_result}
    \vspace{-8pt}
\end{figure*}

\begin{figure*}[t]
  \centering
  \resizebox{0.9\linewidth}{!}{
   \includegraphics{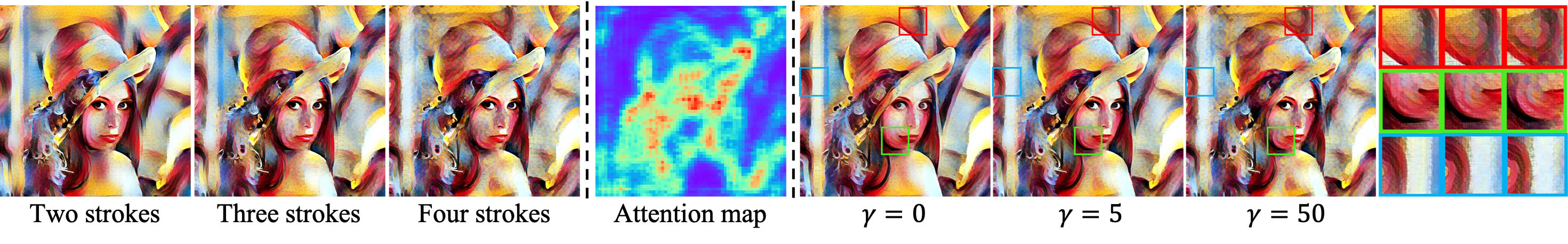} }
   % \vspace{-5pt}
  \caption{Left: fusion strategy with varying number of stroke sizes, Right: level of detail control via regulating the smoothing factor $\gamma$.}
  \label{fig: stroke_control}  
  \vspace{-10pt}
\end{figure*}

\noindent {\bf Ablation Studies.} We discuss in Section~\ref{sec: Proposed Approach} about the contributions of the proposed modules. The self-attention module is responsible for capturing attention characteristics of content image, such that the multi-scale style swap module and the multi-stroke module can work together to integrate multiple stroke patterns. Several evaluations are performed to verify the effectiveness of the coupling framework.

We first train an autoencoder by removing the self-attention module. Since no guidance from the attention map, we apply an average fusion strategy in the multi-stroke fusion module during the testing phase. We name this method as AAMS(-SA). Figure~\ref{fig: ablation} shows the comparison with our method AAMS. Our stylized result with self-attention module demonstrates dramatical improvement on visual effect compared with AAMS(-SA), mainly in two aspects: 1) Our attention-aware method emphasizes the salient regions by painting with fine-grained stroke. 2) The multi-stroke fusion allows abundant integration among stroke patterns and presents discriminative style patterns locally without sacrificing the holistic perception. 

Apart from the cooperation between self-attention and multi-stroke fusion module, the multi-scale style swap also form a compact affiliation with the fusion module. Without the multi-scale style swap module, the style transfer will be restricted to single stroke pattern. We present the final stylized results in Figure~\ref{fig: stroke_result} with different stroke patterns. The stroke size is controlled by changing the scaling coefficient $\beta$ in eq.~\eqref{eq: multi-scale style feature}. From left to right, the larger the stroke size, the coarser stylized pattern emerges. When multiple stroke patterns are generated together, the multi-stroke fusion module can integrate them into one stylized image. 

\subsection{Quantitative Evaluation}

\noindent {\bf User Studies.}
As artistic style transfer is a highly subjective task, we resort to user studies to better evaluate the performance of the proposed method. Since Style-Swap~\cite{chen2016fast} only transfers low-level information, resulting in insufficient stylization effects. We compare the proposed method to the other three competing methods, i.e., AdaIN~\cite{huang2017arbitrary}, WCT~\cite{li2017universal} and Avatar-Net~\cite{sheng2018avatar}. We use 10 content images and 15 style images collected from the aforementioned method and synthesize 150 images for each method, from which we show 20 randomly chosen content and style combinations to each subject.

\begin{figure}[t]
  \centering
  \resizebox{0.9\linewidth}{!}{
   \includegraphics{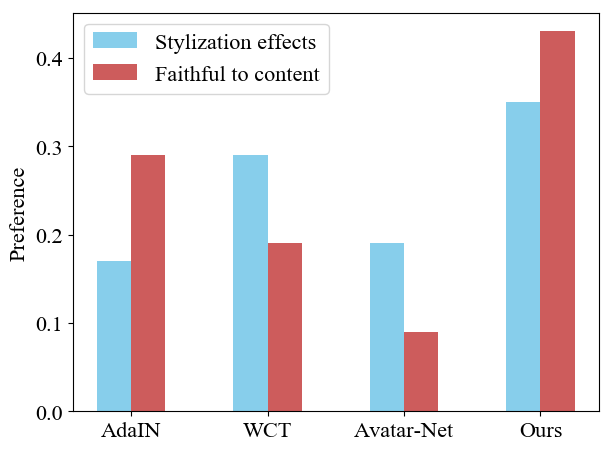} }
  % \vspace{-5pt}
  \caption{User study results.}
  \label{fig: user_study}
    \vspace{-12pt}
\end{figure}

We conduct two user studies on the results in terms of the stylization effects and faithfulness to content characteristics. In the first study, we ask the participants to select which stylized image better transfers both color and texture patterns of style image to content image. In the second study, we ask the participants to select which stylized image better preserves the content characters with less distorted artifacts. In each question, we show the participants the content/style pair and the stylized results of each method in a random order. We finally collect totally 600 votes from 30 subjects and demonstrate the preference results in Figure~\ref{fig: user_study}. The studies show that our method receives most votes for both stylization and attention consistency. The better preservation for visual attention enables perceptual promotion for stylization. 

\noindent{\bf Consistency Evaluation.} To further evaluate the effectiveness on preserving the attention consistency of our method, we propose to adopt saliency metrics~\cite{bylinskii2018different} to measure the saliency similarity between content and stylized image pairs. These metrics are widely used to evaluate a saliency model's ability to predict ground truth human fixations. We generate the saliency maps of stylized images and their corresponding content images using SALGAN~\cite{Pan_2017_SalGAN}, a state-of-the-art saliency prediction method. We then employ five metrics to evaluate the degree of visual attention consistency for 400 content/stylized image pairs. Comparison results between the proposed method and AAMS(-SA) are shown in Table~\ref{tab:consistency}. With our AAMS framework, the attention consistency is dramatically improved for all the evaluation metrics. The results further verify the ability in maintaining saliency for style transfer. 

\noindent {\bf Speed Analysis.} We show the run time performance of our method and prior feed-forward methods in Table~\ref{tab:efficiency}. Results are obtained with a 12G Tesla M40 GPU and averaged over 400 transfers. Among the patch based methods (Style-Swap, Avatar-Net), our method achieves comparable speed even with multi-scale feature processing. Given three strokes, it takes averagely 0.80 and 0.94 seconds to transfer images with size 256 $\times$ 256 and 512 $\times$ 512 respectively. Since the running time increases with stroke numbers in a controllable trend, the tradeoff between efficiency and details diversity should be considered.

\begin{table}[t]
  \centering\scriptsize
  % \vspace{-5pt}
  \resizebox{0.7\linewidth}{!}{
\begin{tabular}{c|c|c}
  \hline
  \hline
  \multirow{2}{*}{\textbf{Method}}&
 \multicolumn{2}{c}{\textbf{Execution Time (sec)}}\\
 \cline{2-3}
& 256$\times$256 & 512$\times$512\\
 \hline
 AdaIN & 0.09 & 0.17 \\
 WCT & 0.92 & 1.05 \\
 Style-Swap & 1.96 & 5.77 \\
 Avatar-Net & 0.78 & 0.86 \\

 \hline
 Ours (2-stroke) & 0.74 & 0.81 \\
 Ours (3-stroke) & 0.80 & 0.94 \\
 Ours (4-stroke) & 0.92 & 1.15 \\ 
 Ours (8-stroke) & 0.94 & 1.18 \\
 \hline
 \hline
\end{tabular}}
    \vspace{5pt}
  \caption{Run time performance. We show runtime test of different stroke numbers with our method and other feedforward methods.}
  \label{tab:efficiency}   
  \vspace{-10pt}
\end{table}

\subsection{Runtime Control}

Given the learnt attention map from the self-attention autoencoder, our method can not only accommodate different requirements from users by providing different controls on the multi-stroke fusion effect, but also achieves automatic spatial stroke size control.

\noindent {\bf Multi-stroke Fusion Control.} One of the advantage towards the prior methods is that our method can effectively integrate multiple stroke patterns with different control strategies, including the number of stroke sizes and level of detail control. As discussed in the previous section, the different scale of high-level style feature will lead to different statistics after a style swap process with the same content feature. Given arbitrary number of stroke sizes, our method allows flexible fusion among these strokes. The left part of Figure \ref{fig: stroke_control} demonstrates three fusion results with different number of stroke sizes. Note that by integrating with more stroke sizes, the stylized result presents more varying patterns with different stroke boundaries, the reason for this is that our clustering algorithm produces different attention distributions for different number of clusters, which results in varying weight maps. The right part of Figure \ref{fig: stroke_control} shows our ability in regulating level of detail control for four-stroke fusion. According to eq.~\eqref{eq: smoothing weight}, the smoothing factor $\gamma$ determines the weight distribution for all stroke sizes. As shown in Figure \ref{fig: stroke_control}, the larger it is, the more discriminative for each stroke within corresponding regions.

\begin{figure}[t]
  \centering
  \resizebox{0.9\linewidth}{!}{
   \includegraphics{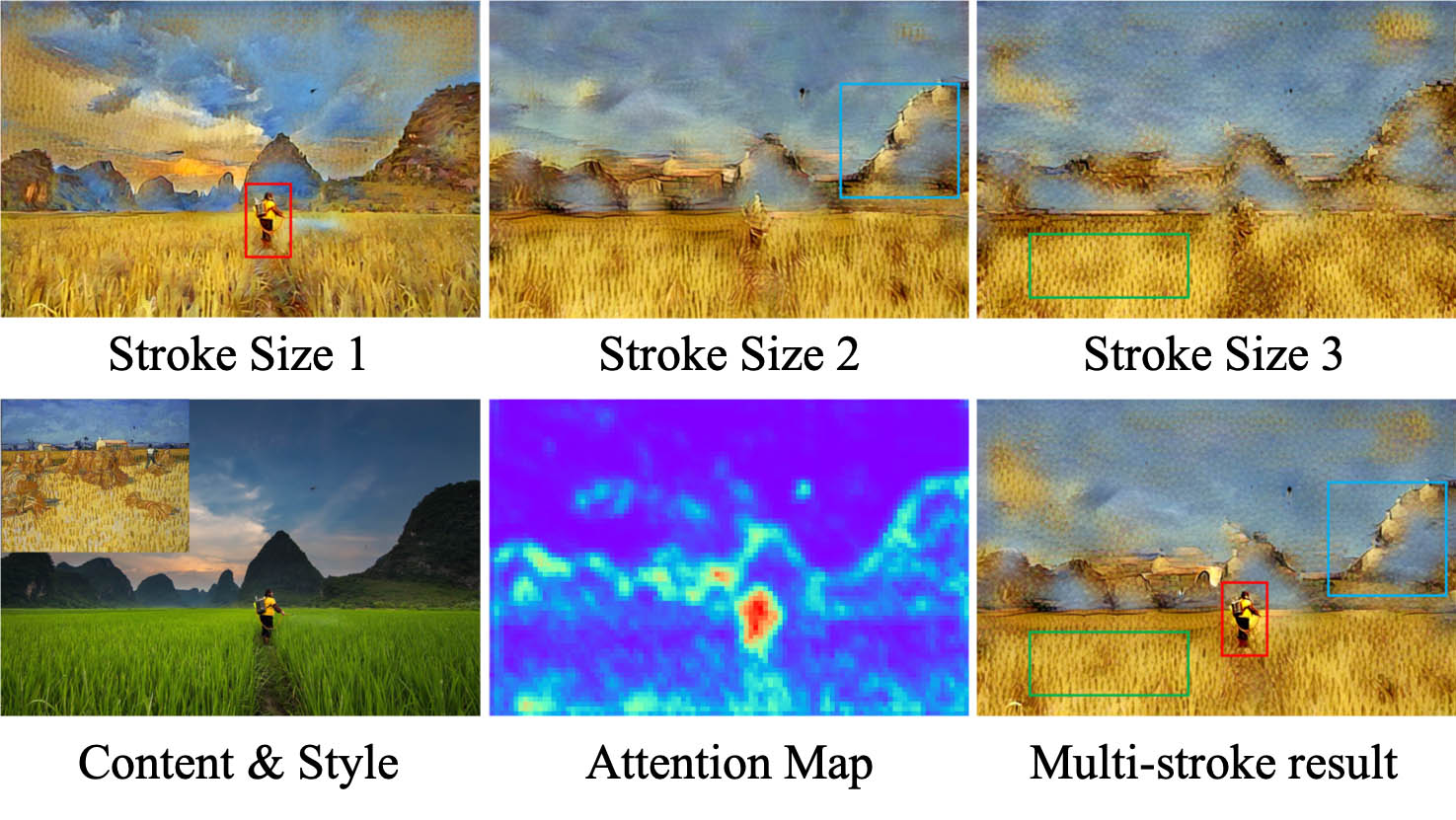} }
  \caption{Our method enables automatic spatial stroke size control.}
  \label{fig: multi-stroke fusion}
    \vspace{-10pt}
\end{figure}

\noindent {\bf Automatic Spatial Stroke Size Control.} Spatial stroke size control refers to properly utilizing strokes of different sizes to render regions with different levels of detail in an image. The gist is that for a visually plausible stylized artwork, we usually hope to use stroke of small size for salient regions (e.g., objects, humans), while large size for less salient ones (e.g., sky, grassland). To this end, previous method \cite{jing2018stroke} generally employs several hand-crafted masks and then stylize corresponding regions separately. Our method, however, empowers spatial control in an automatic way due to the self-attention mechanism. As shown in Figure \ref{fig: multi-stroke fusion}, three stroke sizes are adopted to render regions of different saliency, i.e., the fine stroke patterns on the farmer, the middle stroke patterns on the mountain and coarse stroke patterns on the grass. In particular, these regions are partitioned and integrated automatically under the guidance of content attention map during our multi-stroke fusion procedure, and can be further adjusted by changing the Gaussian variance in attention filter module. By further combining with multiple styles, our method provides automatic solution for spatial control (which refers to transferring each region with different styles in the content image) instead of mask controlling methods \cite{gatys2017controlling,li2017universal}.

\section{Conclusion}
\label{sec: Conclusion}

In this paper, we propose an attention-aware multi-stroke style transfer model for arbitrary styles, which can preserve attention consistency and achieve multi-stroke fusion control and automatic spatial stroke size control in the output result. To accomplish this, we first introduce the self-attention mechanism into the bottleneck layer of an autoencoder framework, then perform a multi-scale style swap to produce multiple swapped features with different stroke sizes. By combining with attention map obtained from content feature, the fusion module can integrate distinct stroke patterns into different regions harmoniously. Experimental results demonstrate the effectiveness of our method in generating favorable results in terms of stylization effects and visual consistency with content image.

%-------------------------------------------------------------------------

%-------------------------------------------------------------------------
\clearpage

{\small
\bibliographystyle{ieee}
\bibliography{aams}
}

\newpage

\section{Appendix}

\subsection{Implementation Details}
We assemble self-attention module into the bottleneck layer of an encoder-decoder framework to form our self-attention autoencoder. Here we present more details of the network architecture.

\subsubsection{Encoder-decoder Architecture}

Table \ref{tab:encoder} and \ref{tab:decoder} illustrate the detailed configurations of the encoder and decoder, respectively. The encoder is made of the first few layers of the VGG-19~\cite{simonyan2014very} network. We take input image with size 512 $\times$ 512 $\times$ 3 as an example and list the feature size for each layer. The max pooling operation is replace by the average pooling operation. The decoder is symmetric to the encoder, with all pooling layers replaced by nearest up-sampling. All convolutional layers use reflection padding to avoid border artifacts~\cite{huang2017arbitrary}. There are some notations; N: the number of output channels, K: kernel size, S: stride size. 

As suggested in~\cite{li2017universal,sheng2018avatar}, it is advantageous to match features across different levels in the VGG-19 encoder to fully capture the charateristics of the style. We use skip connections to perform {}style enhancement using adaptive instance normalization~\cite{huang2017arbitrary}. The three connections are {\em conv}1\_1 $\rightarrow$ {\em inv\_con}1\_2, {\em conv}2\_1 $\rightarrow$ {\em inv\_con}2\_2, {\em conv}3\_1 $\rightarrow$ {\em inv\_con}3\_2, feeding with both output features.

\begin{table*}[t]
  \centering\scriptsize
  % \vspace{-5pt}
  % \captionsetup{font={large}}
  \caption{Details of the encoder. We take input image with size 512 $\times$ 512 $\times$ 3 as an example.}
   \vspace{7pt}
  \resizebox{0.9\linewidth}{!}{
  \begin{tabular}{lll}
 \hline
 Layer &  Layer Information & Feature Size\\
 \hline
 {\em conv}1\_1 & Conv(N64, K3x3, S1), ReLU & (512, 512, 3) $\rightarrow$ (512, 512, 64) \\
 {\em conv}1\_2 & Conv(N64, K3x3, S1), ReLU & (512, 512, 64) $\rightarrow$ (512, 512, 64) \\
 {\em pool}\_1 & AveragePooling(K2x2, S2) & (512, 512, 64) $\rightarrow$ (256, 256, 64) \\

 {\em conv}2\_1 & Conv(N128, K3x3, S1), ReLU & (256, 256, 64) $\rightarrow$ (256, 256, 128) \\
 {\em conv}2\_2 & Conv(N128, K3x3, S1), ReLU & (256, 256, 128) $\rightarrow$ (256, 256, 128) \\
 {\em pool}\_2 & AveragePooling(K2x2, S2) & (256, 256, 128) $\rightarrow$ (128, 128, 128) \\

 {\em conv}3\_1 & Conv(N256, K3x3, S1), ReLU & (128, 128, 128) $\rightarrow$ (128, 128, 256) \\
 {\em conv}3\_2 & Conv(N256, K3x3, S1), ReLU & (128, 128, 256) $\rightarrow$ (128, 128, 256) \\
 {\em conv}3\_3 & Conv(N256, K3x3, S1), ReLU & (128, 128, 256) $\rightarrow$ (128, 128, 256) \\
 {\em conv}3\_4 & Conv(N256, K3x3, S1), ReLU & (128, 128, 256) $\rightarrow$ (128, 128, 256) \\
 {\em pool}\_3 & AveragePooling(K2x2, S2) & (128, 128, 256) $\rightarrow$ (64, 64, 256) \\
 {\em conv}4\_1 & Conv(N512, K3x3, S1), ReLU & (64, 64, 256) $\rightarrow$ (64, 64, 512) \\
 \hline
\end{tabular}}
    % \vspace{5pt}
  \label{tab:encoder}   
  % \vspace{-8pt}
\end{table*}

\begin{table*}[t]
  \centering\scriptsize
  % \vspace{-5pt}
  % \captionsetup{font={large}}
  \caption{Details of the decoder.}
   \vspace{7pt}
  \resizebox{0.9\linewidth}{!}{
  \begin{tabular}{lll}
 \hline
 Layer &  Layer Information & Feature Size\\
 \hline
 {\em inv\_conv}4\_1 & Conv(N256, K3x3, S1), ReLU & (64, 64, 512) $\rightarrow$ (64, 64, 256) \\
 {\em upsample}\_1 & Nearest Upsampling(x2) & (64, 64, 256) $\rightarrow$ (128, 128, 256) \\
 {\em inv\_conv}3\_4 & Conv(N256, K3x3, S1), ReLU & (128, 128, 256) $\rightarrow$ (128, 128, 256) \\
 {\em inv\_conv}3\_3 & Conv(N256, K3x3, S1), ReLU & (128, 128, 256) $\rightarrow$ (128, 128, 256) \\
 {\em inv\_conv}3\_2 & Conv(N256, K3x3, S1), ReLU & (128, 128, 256) $\rightarrow$ (128, 128, 256) \\
 {\em inv\_conv}3\_1 & Conv(N128, K3x3, S1), ReLU & (128, 128, 256) $\rightarrow$ (128, 128, 128) \\
 {\em upsample}\_2 & Nearest Upsampling(x2) & (128, 128, 128) $\rightarrow$ (256, 256, 128) \\
 {\em inv\_conv}2\_2 & Conv(N128, K3x3, S1), ReLU & (256, 256, 128) $\rightarrow$ (256, 256, 128) \\
 {\em inv\_conv}2\_1 & Conv(N64, K3x3, S1), ReLU & (256, 256, 128) $\rightarrow$ (256, 256, 64) \\
 {\em upsample}\_3 & Nearest Upsampling(x2) & (256, 256, 64) $\rightarrow$ (512, 512, 64) \\
 {\em inv\_conv}1\_2 & Conv(N64, K3x3, S1), ReLU & (512, 512, 64) $\rightarrow$ (512, 512, 64) \\
 {\em inv\_conv}1\_1 & Conv(N3, K3x3, S1), ReLU & (512, 512, 64) $\rightarrow$ (512, 512, 3) \\
 \hline
\end{tabular}}
    % \vspace{5pt}
  \label{tab:decoder}   
  % \vspace{-6pt}
\end{table*}

\subsubsection{Self-Attention Module}
\begin{figure*}[t]
  \centering
  \resizebox{0.8\linewidth}{!}{
   \includegraphics{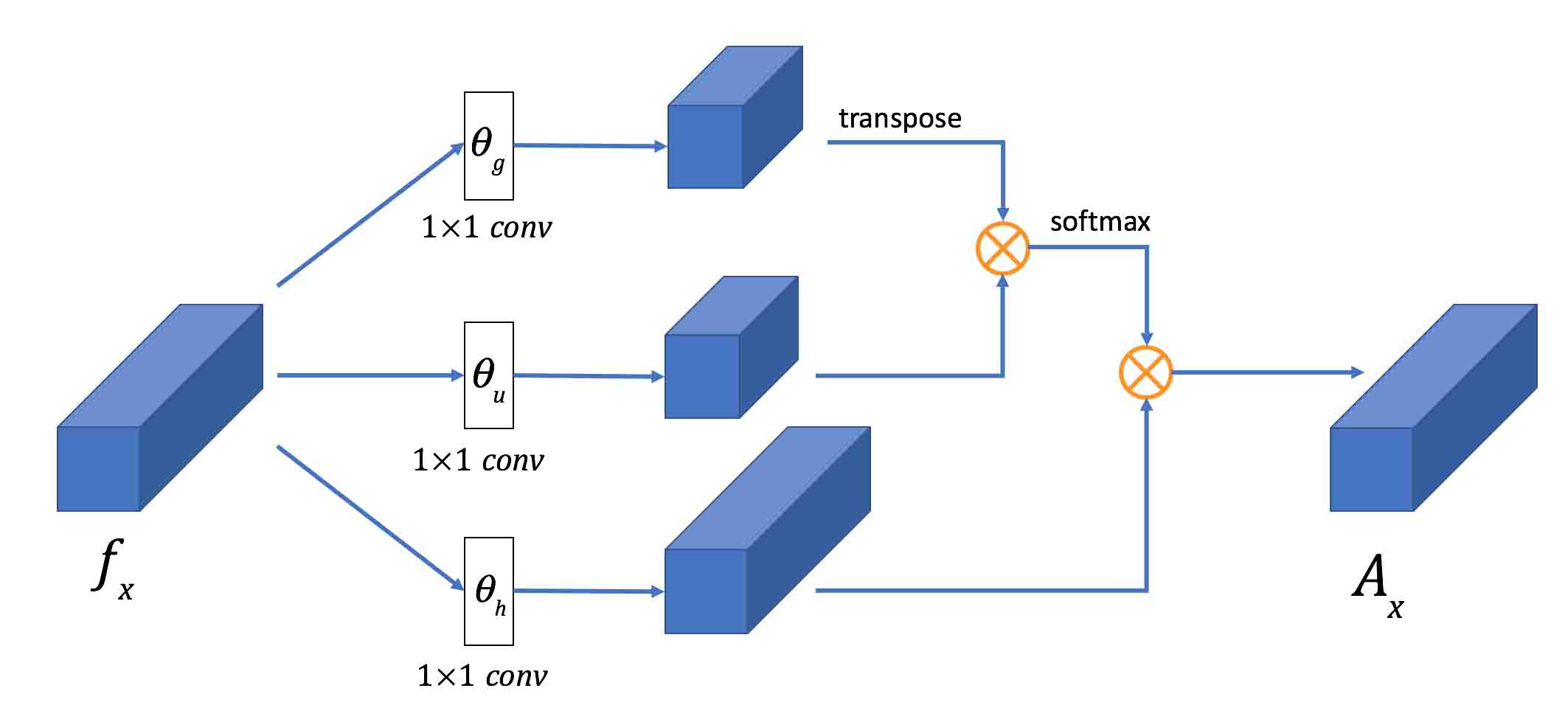} }
   % \vspace{-5pt}
     % \captionsetup{font={large}}
  \caption{The architecture of our self-attention module. The $\otimes$ denotes matrix multiplication.}
  \label{fig: SA_module}  
  % \vspace{-3pt}
\end{figure*}

The architecture of the self-attention module is shown in Figure~\ref{fig: SA_module}. Different from the way used in~\cite{zhang2018self}, where the output of self-attention feature map is added back to the input feature map to learn non-local evidence. We proposed to obtain a self-attention residual $R_{x}$ by multiplying the feature map $f_{x}$ with self-attention feature map $A_{x}$, and find it is effective to capture saliency characteristics.

\subsection{Experiments and Results}
\subsubsection{Extra ablation study}

\begin{figure*}[t]
  \centering
  \resizebox{0.85\linewidth}{!}{
   \includegraphics{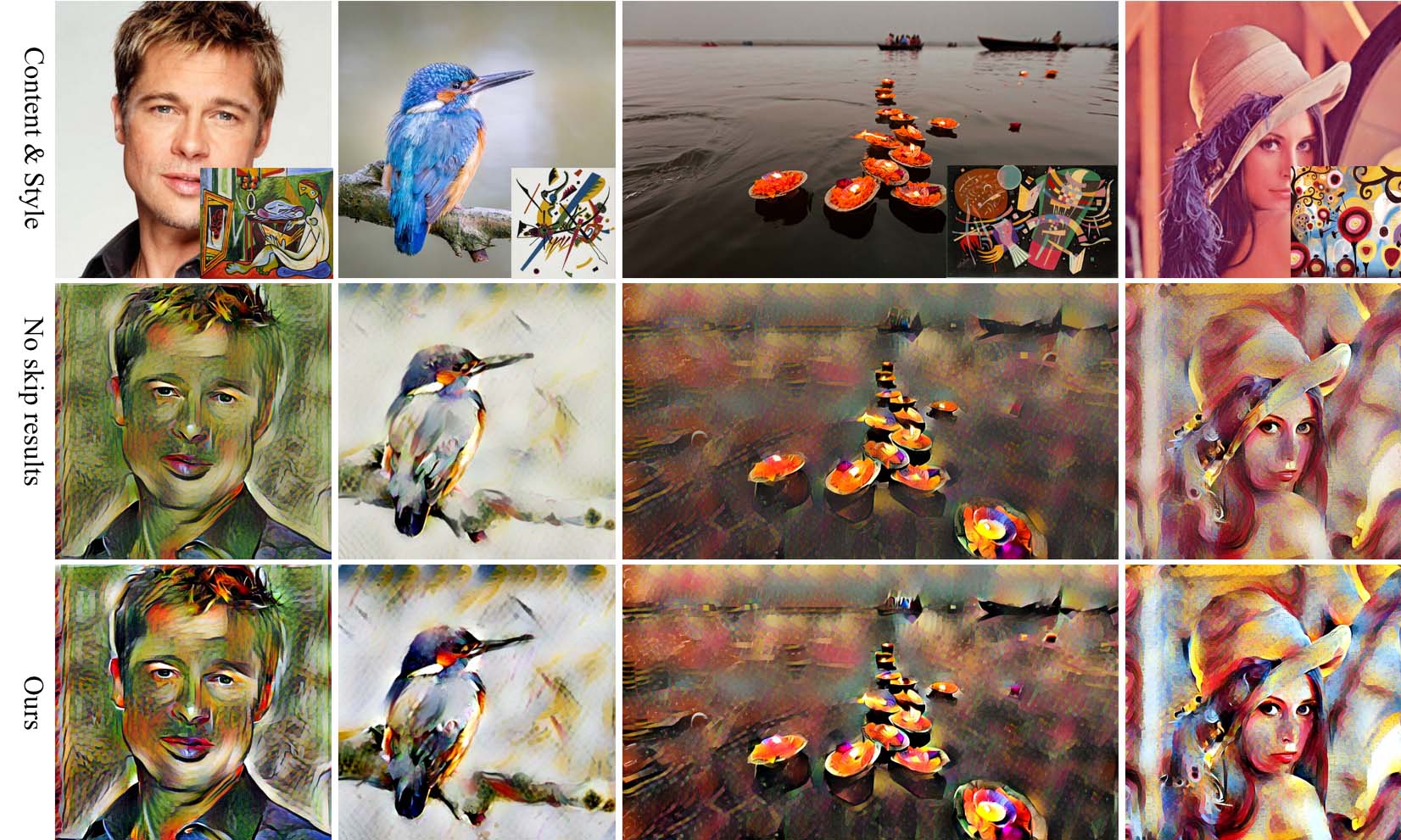} }
   % \vspace{-5pt}
     % \captionsetup{font={large}}
  \caption{The comparison between method without the style enhancement by removing the skip connections. By matching multiple low level features, our results exhibit higher fidelity to styles in terms of color saturation.}
  \label{fig: style_enhance}  
  % \vspace{-5pt}
\end{figure*}

To demonstrate the capability of skip connections for style enhancement. We present the stylized results without the connections in Figure~\ref{fig: style_enhance}. By matching features across multiple levels, the results could capture more low-level characteristics (e.g., colors) of style images, thus exhibit higher fidelity to styles in terms of color saturation.

\subsubsection{More results of our method}

\begin{figure*}[t]
  \centering
  \resizebox{0.9\linewidth}{!}{
   \includegraphics{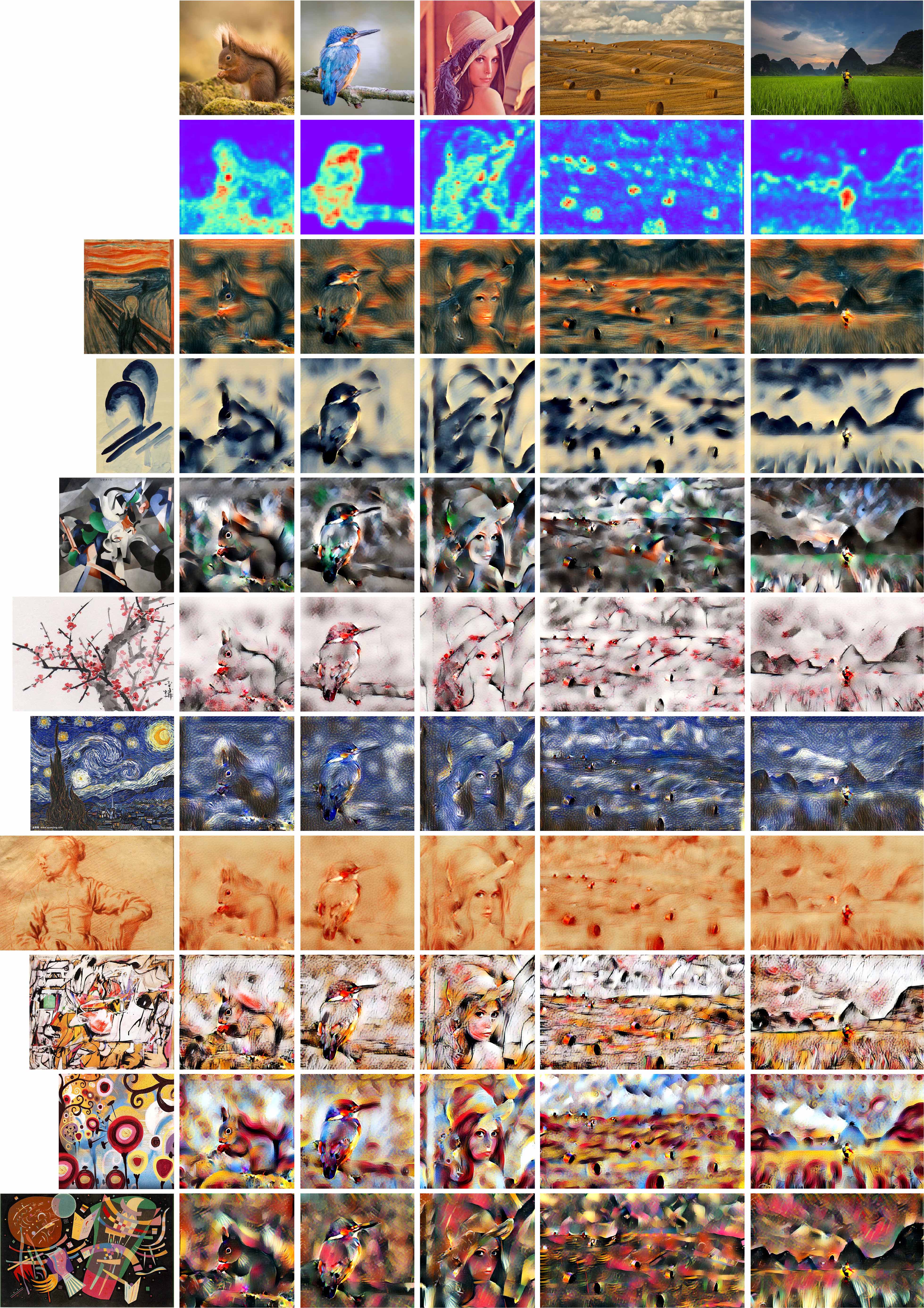} }
   % \vspace{-5pt}
     % \captionsetup{font={large}}
  \caption{More results of our attention-aware multi-stroke style transfer method (Set 1). The {\bf 1st row} is the content images, the {\bf 2nd row} is the corresponding attention maps and the {\bf 1st column} is the style
images.}
  \label{fig: matrixa}  
  % \vspace{-10pt}
\end{figure*}

\begin{figure*}[t]
  \centering
  \resizebox{0.9\linewidth}{!}{
   \includegraphics{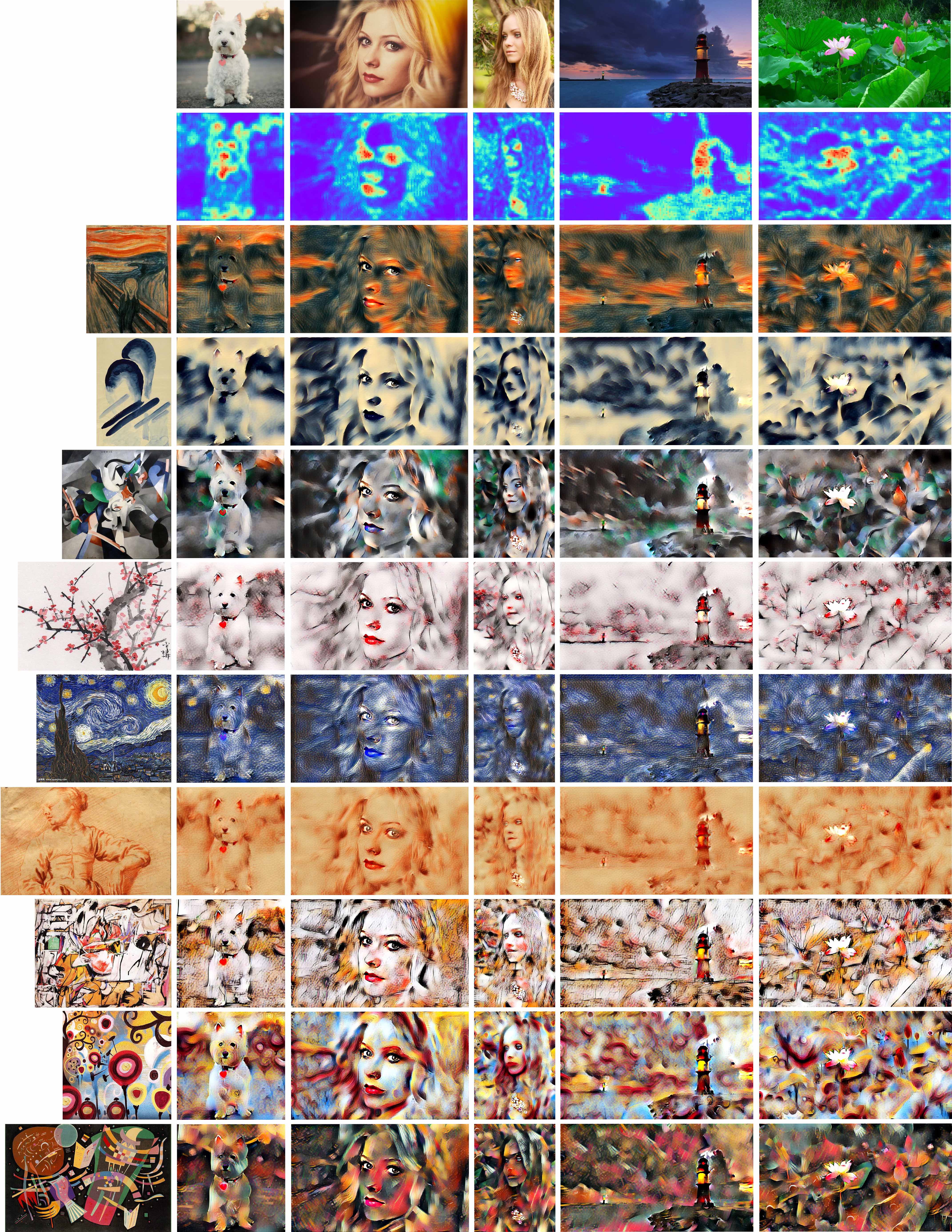} }
   % \vspace{-5pt}
     % \captionsetup{font={large}}
  \caption{More results of our attention-aware multi-stroke style transfer method (Set 2). The {\bf 1st row} is the content images, the {\bf 2nd row} is the corresponding attention maps and the {\bf 1st column} is the style
images.}
  \label{fig: matrixb}  
  % \vspace{-10pt}
\end{figure*}

In this part, we show some additional stylization results by the proposed method, as visualized in Figure~\ref{fig: matrixa} and~\ref{fig: matrixb}. Following the default setting in the main paper, we use three stroke scenario for the proposed style transfer here.

\subsection{Multi-stroke Fusion Control}

Here we explain more details of the advantage of our attention-aware multi-stroke method.

\subsubsection{Stroke control vs weight control}

\begin{figure*}[t]
  \centering
  \resizebox{0.9\linewidth}{!}{
   \includegraphics{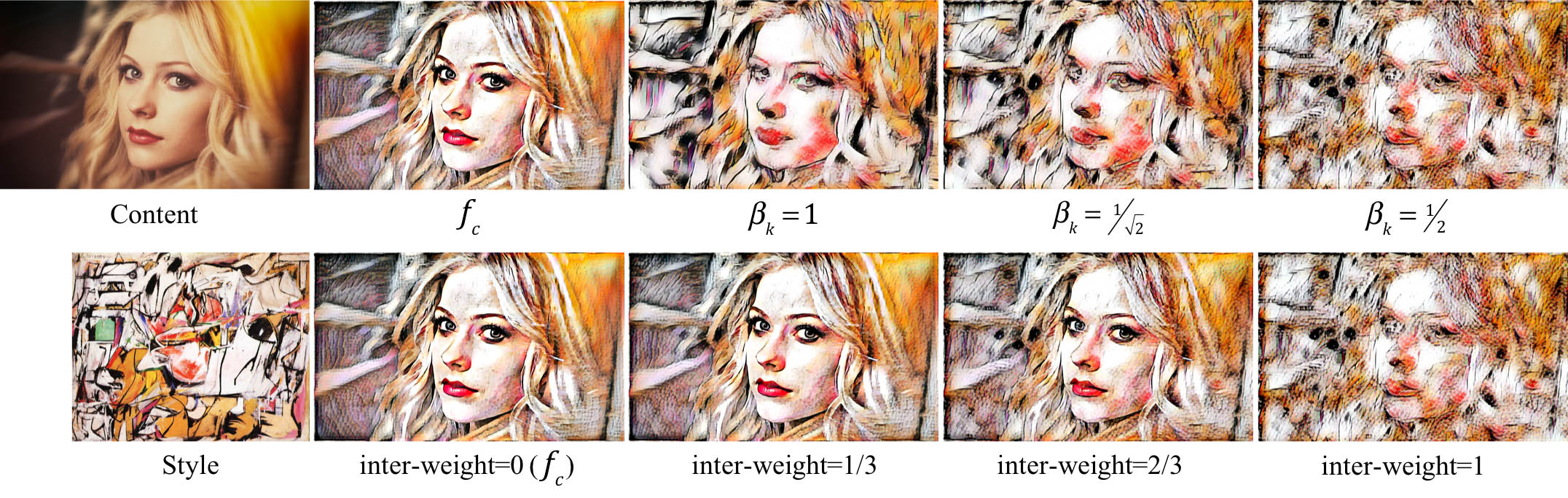} }
   % \vspace{-5pt}
     % \captionsetup{font={large}}
  \caption{Comparison between our multi-scale stroke control and generally used weight control. The weight control simply conduct interpolation as weighted sum of content and stylized features, which demonstrate minor variations among range [0,1]. Our method could flexibly generate continuous and discriminative stylized patterns.}
  \label{fig: stroke_size}  
  % \vspace{-10pt}
\end{figure*}

The weight control refers to controlling the balance between stylization and content preservation. This strategy has been adopted in previous style transfer methods~\cite{huang2017arbitrary,li2017universal,sheng2018avatar}. As visualized in Figure~\ref{fig: stroke_size}, the weight control strategy directly interpolate on deep feature space as weighted sum of content and stylized features, demonstrating minor variations among range [0, 1]. Our multi-scale style swap enables continuous and discriminative stylized patterns by changing the scale coefficient in eq.~\eqref{eq: multi-scale style feature} of the paper, and further generate integrated results via different combinations efficiently.

\subsubsection{Fusion control strategy}
\begin{figure*}[t]
  \centering
  \resizebox{0.9\linewidth}{!}{
   \includegraphics{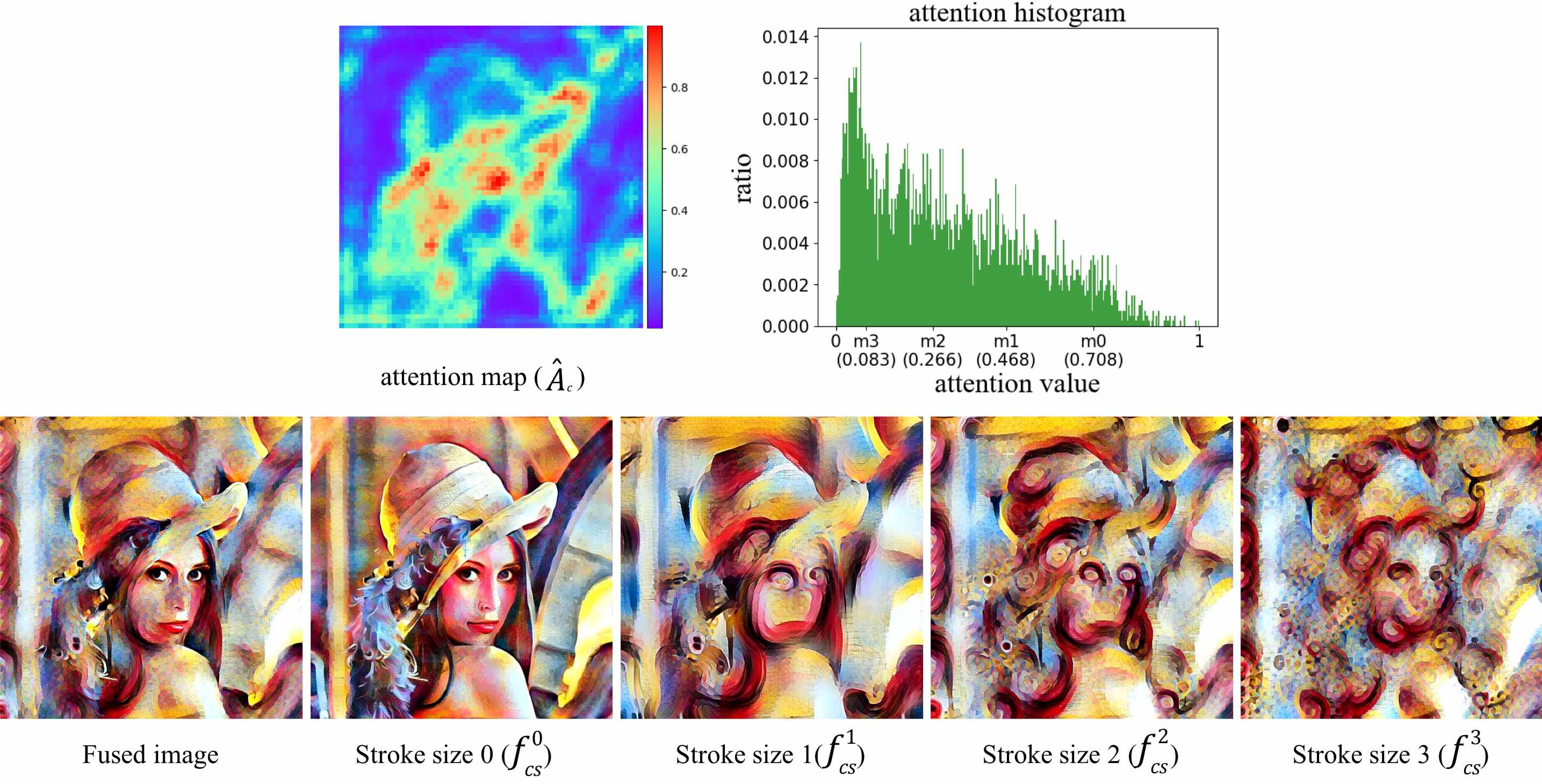} }
   % \vspace{-5pt}
     % \captionsetup{font={large}}
  \caption{The procedure for our multi-stroke fusion. In the attention histogram, we mark the attention value of clustering centers obtained by applying k-means on the attention map $\hat{A}_{c}$, with higher attention values assigned to finer stroke patterns. The integrated feature map is generated seamlessly as the weighted sum of these stroke feature maps according to the proposed fusion strategy.}
  \label{fig: attention_mean}  
  % \vspace{-10pt}
\end{figure*}

As mentioned in Section 4.4, our method can effectively integrate multiple stroke patterns with different control strategies. We present the fusion procedure in Figure~\ref{fig: attention_mean}. Given $\emph{K}$+1 stroke feature maps $(f_{cs}^{0}, f_{cs}^{1}, \ldots, f_{cs}^{K})$ and the corresponding attention map $\hat{A}_{c}$, we first generate $\emph{K}$+1 clustering centers with attention values according to eq.~\eqref{eq: kmeans}, and then assign sequentially for stroke sizes, with higher attention values for finer stroke patterns. The integrated feature map is the weighted sum of the $\emph{K}$+1 stroke feature maps based on eq.~(\ref{eq: Multi-stroke Fusion}-\ref{eq: smoothing weight}) in the paper.

\begin{figure*}[t]
  \centering
  \resizebox{0.9\linewidth}{!}{
   \includegraphics{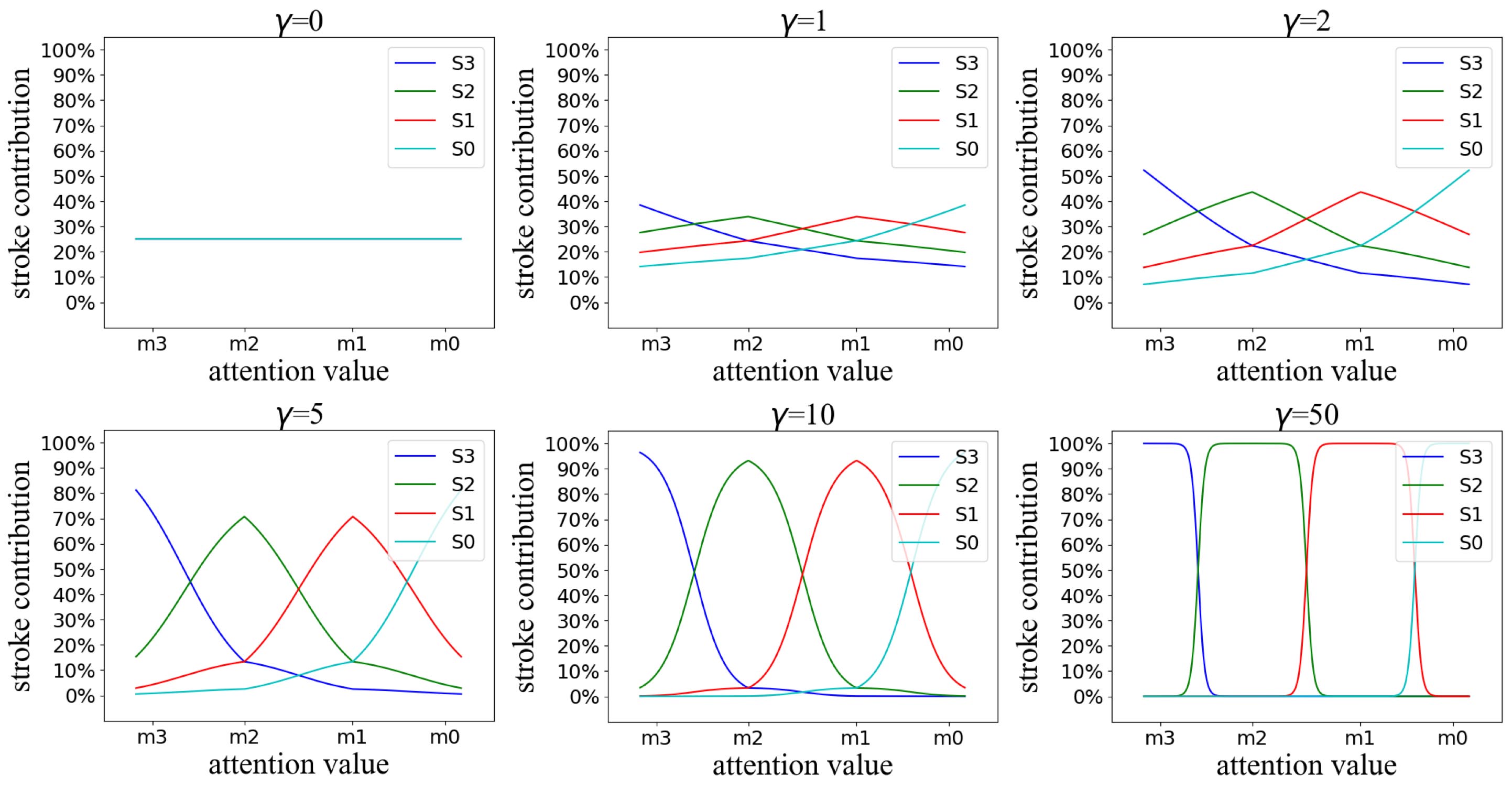} }
   % \vspace{-5pt}
     % \captionsetup{font={large}}
  \caption{Varying values of smoothing factor $\gamma$ affecting the fusion contribution for each stroke pattern. Each stroke size is assigned with an intensity center for a specific attention area. With the increasing of $\gamma$, each stroke pattern tends to contribute more on its own attention area, leading to a more discriminative effect among these stroke patterns.}
  \label{fig: gamma_control}  
  % \vspace{-10pt}
\end{figure*}

The level of detail for multi-stroke fusion can be controlled by the smoothing factor $\gamma$. As visualized in Figure~\ref{fig: gamma_control}, we present the influence of different smoothing values for four-stroke fusion scenario. The larger it is, the more contributions for single stroke on corresponding attention area, leading to a more discriminative effect among these stroke patterns.

\end{document}